\documentclass[10pt,twocolumn,letterpaper]{article}

\usepackage{ijcb}
\usepackage{times}
\usepackage{epsfig}
\usepackage{graphicx}
\usepackage{amsmath}
\usepackage{amssymb}
\usepackage{cite}
\usepackage{multirow}

\usepackage{caption}
\usepackage{subcaption}
\usepackage{setspace}
\usepackage{soul}

\usepackage[pagebackref=true,breaklinks=true,letterpaper=true,colorlinks,bookmarks=false]{hyperref}

\ijcbfinalcopy % *** Uncomment this line for the final submission

\ifijcbfinal\fi

\makeatletter
\def\ps@IEEEtitlepagestyle{
\def\@oddfoot{\mycopyrightnotice}
\def\@evenfoot{}
}
\def\mycopyrightnotice{
{
    Original title: \textit{Unconstrained Face Identification using Ensembles trained on Clustered Data}
    \hfill \footnotesize 978-1-7281-9186-7/20/\$31.00 \copyright 2020 IEEE\hfill}
}
\makeatother

\newcommand{\ackfootnote}[1]
{
  \begingroup
  \renewcommand\thefootnote{}\footnote{#1}%
  \addtocounter{footnote}{-1}%
  \endgroup
}

\begin{document}

    \title{
        \large
        \vspace{-2.0ex}
        [\st{Unconstrained}] Open-set Face Recognition using Ensembles trained on Clustered Data
        \vspace{-1.5ex}
    }
    \author{
        \normalsize Rafael Henrique Vareto and William Robson Schwartz\\
        \normalsize Smart Sense Laboratory, Department of Computer Science\\
        \normalsize Universidade Federal de Minas Gerais, Belo Horizonte, Brazil\\
        {\tt\small \{rafaelvareto, william\}@dcc.ufmg.br}\\
    }

   \maketitle
   \thispagestyle{empty}

    \begin{abstract}
	\vspace{-1.0em}
    Open-set face recognition describes a scenario where unknown subjects, unseen during training stage, appear on test time. 
    Not only it requires methods that accurately identify individuals of interest, but also demands approaches that effectively deal with unfamiliar faces.
    This work details a scalable open-set face identification approach to galleries composed of hundreds and thousands of subjects. 
    It is composed of clustering and ensemble of binary learning algorithms that estimates when query face samples belong to the face gallery and then retrieves their correct identity.
    The approach selects the most suitable gallery subjects and use the ensemble to improve prediction performance.
    We carry out experiments on well-known \textsc{lfw} and \textsc{ytf} benchmarks.
    Results show that competitive performance can be achieved even when targeting scalability.
    \vspace{-1.0em}
\end{abstract}
    \ackfootnote{\textsc{doi}: \href{https://ieeexplore.ieee.org/document/9304882}{10.1109/IJCB48548.2020.9304882}}
    % \ackfootnote{\mycopyrightnotice}

   \section{Introduction}
\label{sec:introduction}

% Intro
Biometrics is used to identify and authenticate individuals using a set of discriminative physical and behavioral characteristics, which is inherent to every person. 
Few biometric traits are simultaneously non-invasive and accurate like face recognition.
In fact, face recognition has been in use for decades and persists as one of the most widely used biometric traits.

% Problem
Most face recognition approaches analyze facial geometries and their similarity to one another as they attempt to correctly identify a subject against a list of previously registered individuals (face gallery), containing faces with similar characteristics.
However, face recognition is a broad term, commonly used to indicate three closely-related subtasks~\cite{chellappa2010face}: 
\textit{face verification}, which focuses on determining whether a pair of different face images corresponds to the same subject; 
\textit{closed-set face identification}, a task that presents a query image against a number of previously cataloged faces, ensuring that the probe face always contains a corresponding identity in the gallery set; 
and \textit{open-set face identification}, which is similar to closed-set identification with the distinction that it does not guarantee that all query subjects are registered in the face gallery. 

Both closed and open-set face identification tasks can be described as a $1:N$ matching problem, where $N$ equals the number of subjects enrolled in the gallery set.
The former has training and testing data drawn from uniform label and feature spaces whereas the latter comes across situations in which unseen individuals emerge unexpectedly.
More precisely, along with finding out which identity from the face gallery best matches an unknown face sample, open-set face identification systems first have to check whether the probe image indeed belongs to any of the registered subjects. 
In summary, open-set approaches behave like closed-set identification for known individuals and also label non-identified persons with an ``unknown subject'' category.

% Motivation
There is a great demand for open-set technology as the vast majority of real-world identification problems consist of a finite number of persons of interest in comparison to innumerable unknown individuals. 
As a clear illustration, think of an identification application for law enforcement agencies where lawbreakers' identities are doubtless of interest; on the other hand, an infinite number of law-abiding citizens are not of concern.
For that reason, open-set algorithms should dismiss all unwanted subjects and focus on identifying potential suspects only. 
Still, most researchers have left open-set problems aside and channeled their efforts into closed-set face identification problems despite the real-world appeal~\cite{klare2012heterogeneous, yi2013towards, kan2014stacked, yi2015shared, liu2017sphereface}.
Thusly, there is yet a lot to improve when it comes to open-set face identification.
% ~\cite{klare2012heterogeneous, yi2013towards, kan2014stacked, yi2015shared, dos2016partial, liu2017sphereface}

% Method overview
% The proposed approach incorporates efficient and straightforward techniques to handle subjects not seen during training time. 
The proposed approach incorporates efficient and straightforward techniques, namely Affinity Propagation Clustering (\textsc{apc}) algorithm~\cite{frey2007clustering} and an ensemble of Partial Least Squares (\textsc{pls})~\cite{rosipal2005overview} models. 
We follow Breiman~\etal's~\cite{breiman1996heuristics} statement that the building of multiple learning algorithms, trained on randomly generated training sets, tends to achieve better predictive performance than the composing classifier alone. 
This approach is inspired by the work of Vareto~\etal\cite{vareto2017towards}, which aggregates each model's response as it sets up a vote-list histogram where each bin comprises a gallery-registered subject.
But differently, for a given query image, the proposed method performs affinity propagation clustering of all gallery subjects and elects the most suitable clusters to train a binary ensemble of classifiers on prominent subjects' face samples only.

% Hypothesis
We believe that vote-list histograms react in a distinctive manner when query face samples correspond to identities previously registered in the gallery set, but present a uniform behavior when probe face images have no corresponding match. 
In fact, we hypothesize that when a probe sample is known, most classifiers would vote for the correct identity or otherwise distribute the votes among distinct gallery-registered individuals.

% Technical advantages
Unlike \textit{one-against-all} learning schemes where generally unbalanced classification models are learned for each person of interest~\cite{huang2011extreme}, the proposed approach is not directly dependent on the number of known subjects since clustering the gallery individuals and selecting the top $k$ clusters significantly reduce the computational time.
Actually, the method consists of balanced data splits.
Qu~\etal\cite{qu2010asymmetric} show that unbalanced class distributions may cause conventional classifiers, such as support vector machines~\cite{steinwart2008support}, to treat a few-sample class like noises and push the classification boundaries so that it may benefit the majority class, culminating in a precision drop of the outnumbered class.
Providentially, the way the proposed method splits data guarantees symmetrical division between positive and negative classes for every model belonging to the ensemble.

% Objectives
% This work details a scalable open-set face identification approach to galleries with hundreds and thousands of subjects. 
To address complexity issues intrinsic to open-set recognition problems, we follow different literature protocols designed for open-set face evaluation on two well-known datasets: Labeled Faces in the Wild (\textsc{lfw})~\cite{huang2008labeled} and Youtube Faces (\textsc{ytf})~\cite{wolf2011face}.
We also check whether associating clustering and learning algorithms as an ensemble is an adequate form of outperforming more complex models.
Then, we evaluate how the method's performance responds to variable ensemble sizes and investigate whether it is possible to offer a trade-off between accuracy and simplicity.
According to experimental results, our approach reports competitive matching accuracy in comparison with other state-of-the-art works on \textsc{lfw} and \textsc{ytf} databases.

% Contributions
The main contributions of this work are:
\emph{(i)} a scalable open-set approach, dismissing the need of retraining with the addition of new individuals; %capable of dealing with high-dimensional feature vectors;
\emph{(ii)} an easy-to-implement method with few trade-off parameters to be estimated;
\emph{(iii)} an algorithm that reduces the training space and learns small ensembles to improve prediction performance;
\emph{(iv)} a comprehensive discussion on the carried-out experiments, considering different protocols;
\emph{(v)} the development of a baseline that can be considered in future works regarding open-set face recognition.  

   \section{Related Works}
\label{sec:related}

Contemporary years have witnessed a significant development on the face recognition field~\cite{wright2009robust, best2014unconstrained, crosswhite2018template}. 
Although IBM stated approximately 50 years ago that humans could be recognized at a computer terminal by something they carry or by a personal characteristic~\cite{jain2008handbook}, only recently has open-set recognition been explored in the literature.
On the other hand, there have been more works intended to solve closed-set and verification problems, either in unrestrained scenarios or in relatively small datasets \cite{tan2010enhanced, zhu2015discriminative, wang2018additive}. 
% Therefore, this section presents an overview regarding the most recent works targeting the open-set face identification task. 
% \cite{wright2009robust, guo2011face, best2014unconstrained, crosswhite2018template}
% \cite{tan2010enhanced, zhu2015discriminative, vareto2017verification, wang2018additive}

% Non-LFW approaches: FRGC, PubFig83, FERET, Yale, Rice, PIE-CMU, NRL datasets
    % [20] 
    % F. Li and H. Wechsler. 
    % Open set face recognition using transduction. 
    % Transactions on Pattern Analysis and Machine Intelligence (TPAMI), 27(11), 2005. 2, 3

    % Kamgar-Parsi, B., Lawson, W., & Kamgar-Parsi, B. (2011). 
    % Toward development of a face recognition system for watchlist surveillance. 
    % IEEE Transactions on Pattern Analysis and Machine Intelligence, 33(10), 1925-1937.

    % dos Santos Junior, C. E., & Schwartz, W. R. (2014, August). 
    % Extending face identification to open-set face recognition. 
    % In 2014 27th SIBGRAPI Conference on Graphics, Patterns and Images (pp. 188-195). IEEE.

    % Yu, H., Fan, Y., Chen, K., Yan, H., Lu, X., Liu, J., & Xie, D. (2019). 
    % Unknown Identity Rejection Loss: Utilizing Unlabeled Data for Face Recognition. 
    % In Proceedings of the IEEE International Conference on Computer Vision Workshops (pp. 0-0).

    Li~\etal\cite{li2005open} developed a transductive inference that includes an option for rejecting unknown subjects. 
    % It behaves like an outlier detection and takes advantage of an automatic threshold selection.
    Kamgar~\etal\cite{kamgar2011toward} identified decision regions in the high-dimensional face space by generating two large sets of borderline images through morphing techniques.
    % Borderline data is developed through morphing along segments of the manifold where data is sparse.
    Santos~\etal\cite{dos2014extending} came up with five approaches, such as commonplace attributes among registered subjects, margin separation or distribution patterns between identification responses.
    Vareto~\etal\cite{vareto2017towards} proposed \textsc{hpls}, assuming that vote-list histograms behave differently for queries comprising gallery-enrolled subjects and unknown individuals as histograms present highlighted bins when the query image matches an identity of interest.
    Yu~\etal\cite{yu2019unknown} designed a new loss function that utilizes unlabeled data to further enhance the discriminativeness of the learned feature representation.

    % The works mentioned above explored a variety of datasets and established different open-set protocols for each one of them.
    The works mentioned above established different open-set protocols for a variety of datasets.
    This forges a problematic situation where it is almost impossible to make legitimate comparisons of research works since they do not follow a mainstream experimental evaluation.
    Some of these methods also overlook the identification stage as they only estimate whether a query sample belongs to the gallery set, without returning its actual identity.
    All these limitations raise the following question: ``What is the most accurate algorithm for open-set face problems under certain shared conditions?''.
    As a result, the following literature works adopted either \textsc{lfw}~\cite{huang2008labeled} or \textsc{ytf}~\cite{wolf2011face} datasets and devised new protocols for open-set assessment.

% LFW
    % [4] 
    % L. Best-Rowden, H. Han, C. Otto, B. F. Klare, and A. K. Jain.
    % Unconstrained face recognition: Identifying a person of interest from a media collection. 
    % Transactions on Information Forensics and Security (TIFS), 9(12), 2014. 2, 3, 6

    % [21] 
    % S. Liao, Z. Lei, D. Yi, and S. Z. Li. 
    % A benchmark study of large-scale unconstrained face recognition. 
    % In International Joint Conference on Biometrics (IJCB). IEEE, 2014. 2, 3

    % [35]
    % Y. Sun, D. Liang, X. Wang, and X. Tang. 
    % Deepid3: Face recognition with very deep neural networks. 
    % arXiv preprint arXiv:1502.00873, 2015. 2

    % Gunther, M., Cruz, S., Rudd, E. M., & Boult, T. E. (2017). 
    % Toward open-set face recognition. 
    % Conference on Computer Vision and Pattern Recognition Workshops (pp. 71-80).
    
    Best-Rowden~\etal\cite{best2014unconstrained} boosted the likelihood of identifying subjects by fusing several media types.
    Liao~\etal\cite{liao2014benchmark} evaluated seven learning algorithms taking as input three different feature descriptors.
    Sun~\etal\cite{sun2015deepid3} designed two deep neural networks with prefix \textsc{DeepID} to learn a richer pool of facial features.
    Martinez~\etal~\cite{martindez2019shufflefacenet} propose a lightweight architecture, \textsc{ShuffleFace}, that use less than 4.5 million parameters.
    % G{\"u}nther~\etal\cite{gunther2017toward} introduced the concepts of known unknowns (known, but uninteresting persons) and unknown unknowns (people never seen before) individuals as they compare three similarity-assessing algorithms in deep feature spaces: cosine similarity, linear discriminant analysis, and extreme value machine~\cite{rudd2017extreme}.
    G{\"u}nther~\etal\cite{gunther2017toward} introduced \textit{known unknowns} (uninteresting persons) and \textit{unknown unknowns} (people never seen before) individuals as they compare three similarity-assessing algorithms in deep feature spaces: Cosine similarity (\textsc{cos}), Linear Discriminant Analysis (\textsc{lda}), and Extreme Value Machine (\textsc{evm})~\cite{rudd2017extreme}.
     
    Despite the fact that the aforementioned approaches attained significant progress in the test-subject search, they still face scalability problems in cases of restricted training time or excessive data.
    The low computational time performance is probably due to the fact that these approaches still present linear asymptotic complexity with the gallery size.
    In such situations, unsatisfactory performance is observed when there are numerous individuals in the face gallery.
    
\begin{figure*}[!ht]
    \vspace{-2.0mm}
    \centering
    \includegraphics[width=0.97\linewidth]{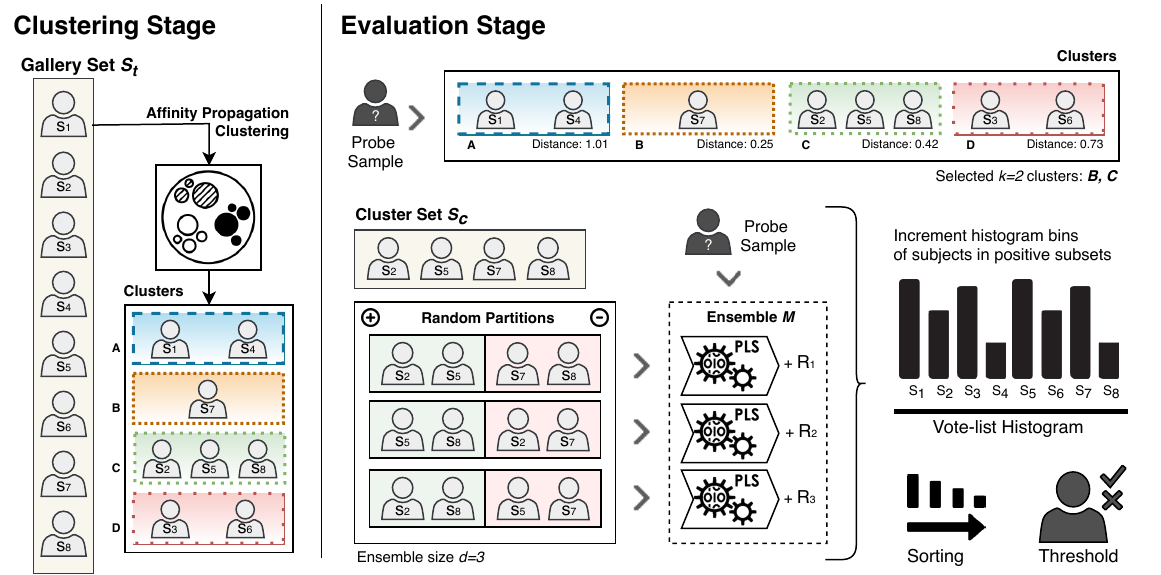}
    \vspace{-2.0mm}
    \caption{A pipeline illustration of proposed approach having an ensemble of classifiers with size $d = 3$ and selecting best $k = 2$ clusters. \textit{\textbf{Clustering:}} Training data $S_{t}$ is initially partitioned into \textsc{apc}-generated subgroups. Then, when a query sample $q$ is presented to the collection of clusters, \textsc{apc} algorithm picks the most similar clusters ($B$ and $C$) and sets up a training subset $S_{c}$. \textit{\textbf{Online Training:}} Training subjects from $S_{c}$ are randomly partitioned into $d$ positive and negative subsets. Random partitions feed \textsc{pls} models in the interest of training the ensemble $M$. \textit{\textbf{Testing:}} The very same query sample $q$ is presented to the ensemble in pursuance of response values $r_{i} \in R$. These values are used to increment the histogram bins of individuals $s_{j} \in S$ randomly sampled in the positive subset (green box). The histogram is sorted in descending order and becomes a ranking. The thresholding is employed to find out whether the probe sample belongs to the face gallery.}
    \label{fig:train-test}
    \vspace{-4.0mm}
\end{figure*}
    
% Open-set Datasets
Few literature datasets introduce open-set face identification protocols.
In fact, \textsc{ijb-a}~\cite{klare2015pushing} seems to be the first open-set benchmark containing ``in-the-wild'' image and video samples.
% It is made up of 500 individuals, 5,712 images, 2,085 videos and 10 random training and testing splits.
% According to \textsc{ijb-a} guidelines, approximately 11\% out of the total number of subjects available are randomly removed from the gallery set in each split.
% Whitelam~\etal\cite{whitelam2017iarpa} has compiled \textsc{ijb-b} in order to solve the problem of having a low number of possible unknown-subject matching as it is up to 1,845 subjects and nearly 22,8 and seven thousand still and motion pictures, respectively.
Recently, Maze~\etal\cite{maze2018iarpa} designed \textsc{ijb-c} on the lookout for evaluating end-to-end systems, that is, algorithms that jointly detect faces and recognize people.
\textsc{ijb-c} includes a total of 31,334 images and 117,542 frames from 11,779 videos.
The three benchmarks are produced by \textsc{iarpa's} Janus program, which in 2019, decided to no longer distribute the datasets for an undisclosed period.
Similarly, \textsc{uccs's} open-set surveillance dataset and respective challenge~\cite{gunther2017unconstrained} are temporarily suspended due to metadata rectification for the third competition edition.

% Datasets unavailability
Recent open-set works dismiss \textsc{iarpa}'s datasets as some investigators claim they contain missing data annotations, low-quality images and videos~\cite{sun2015deepid3, gunther2017toward}.
G{\"u}nther~\etal\cite{gunther2017toward} declare that several issues must be worked out before researchers are able to correctly address the open-set problem.
Some investigators have adopted \textsc{lfw}/\textsc{ytf} as the ``official'' benchmark for open-set face tasks due to the novel protocols released in subsequent research works~\cite{liao2014benchmark, gunther2017toward, martinez2018toward}. 
They intend to come up against the lack of consensus of open-set biometrics.
In the experiments, we evaluate the proposed method on \textsc{lfw} and \textsc{ytf} datasets as we adjust it to meet already-explored open-set protocols.

   \section{Proposed Method}
\label{sec:method}

Noisy training sets may mislead a method's identification and the adoption of subsets of big data is a manner of removing meaningless information.
Clustering reduces the computation time as it avoids comparing a probe image to all training classes, especially when a gallery set is large. 
The designed method, illustrated in Figure~\ref{fig:train-test}, is made of the Affinity Propagation Clustering (\textsc{apc})~\cite{frey2007clustering} algorithm and an ensemble of binary Partial Least Squares (\textsc{pls})~\cite{rosipal2005overview}.

Gallery-composing subjects are submitted to a clustering algorithm, building the collection of clusters $C$.
Given a query face sample $q$, individuals from top $k$ clusters that most resemble $q$ are singled out and their corresponding image samples are employed to learn $d$ classifiers $m \in M^{d}$.
Then, probe $q$ is projected onto the embedding in favor of determining whether its identity lies in the gallery set or comprise an unknown subject.
In fact, probe sample $q$'s corresponding feature descriptor is presented to each model $m$ in search of its response value $r$.
Each value $r_{i} \in R^{d}$ is added to a vote-list histogram of size $|S_{t}|$ as it increments the bins of those subjects in the $i$-th positive subset of model $m_{i} \in M^{d}$.
If the algorithm establishes that a probe image corresponds to an enrolled identity, the vote-list histogram turns into a list of candidates~\cite{vareto2017towards}.

\subsection{Affinity Propagation Clustering}
Affinity Propagation is a centroid-based clustering algorithm that automatically chooses the number of clusters.
Not only does it group facial data samples into different clusters, but it also estimates a representative sample (exemplar) for each one of them.
In our context, it takes in a collection of real-valued similarities between face samples, where the similarity function $p(s_{i},s_{j})$ indicates how well a face image from subject $s_{j}$ is suitable to be an exemplar for subject $s_{i}$.
\textsc{apc} takes as input facial features and identifies exemplars based on certain norms, such as instances similarity and availability.
It iteratively searches for clustering arrangements that satisfy its optimization criterion and stops when cluster boundaries stand consistent over a number of iterations.

\vspace{1.5mm} \noindent\textbf{Clusters Selection.}
The subjects $s \in S_{t}$ available in the gallery set input the \textsc{apc} procedure, resulting in the collection of clusters $C$.
Whenever a probe face sample $q$ turns up, the proposed approach projects $q$ onto the clustered data $c \in C$, comparing it to the estimated representative exemplars (centroids).
The first $k$ clusters from collection $C$, containing individuals that most resemble query sample $q$, are maintained for ensemble learning as they constitute a smaller training set called $S_{c}$.

Container $S_{c}$ contains a list of subject classes that favors high \textit{interclass} similarity since the clustering strategy facilitates the picking of ``look-alike'' individuals.
It eliminates most undesirable subjects and, consequently, endorses the learning of more robust classifiers as it involves distinguishing among visually-related samples.
According to Caruana~\etal\cite{caruana2004ensemble}, the building of more discriminating models allows the method to be optimized with respect to performance metrics such as accuracy, cross entropy and mean precision to mention a few.

% \subsection{PLS Emsemble and Hamming Embedding}
\subsection{Partial Least Squares Ensemble}
The proposed ensemble is an adaptation of a technique called Bootstrap Aggregating~\cite{breiman1996heuristics}, well-known for the acronym \emph{bagging}.
% Algorithms such as neural networks, support vectors, classification and regression trees commonly face instability issues when dealing with unbalanced training data.
Several machine learning algorithms, like \textsc{cnn}, face instability issues when dealing with unbalanced training data.
Favorable results can be attained when bagging is employed since the adoption of multiple learning algorithms tends to accomplish better predictive accuracy than standalone regression and classification models.

\vspace{1.5mm} \noindent\textbf{Online Training.}
% Most bagging techniques consist in sampling from $S_{c}$ uniformly and with replacement, generating $d$ smaller training subsets in which some subject samples may be repeated.
Most bagging techniques consist of uniformly sampling from a face dataset with replacement, generating smaller training subsets in which some subject samples may be repeated. 
However, the developed method employs $d$ random fifty-fifty disjoint splits: each random partitioning encompasses all individuals available in cluster set $S_{c}$ by allocating half of them in the positive and the other half in the negative subset.
The algorithm guarantees a complete utilization of container $S_{c}$, ensures balanced data division, and implicitly provides each known subject with a binary string of dimension $d$.
% \footnote{The association of a subject to positive or negative targets consists in sampling from a Bernoulli distribution with probability $p = 0.5$.}

The $d$ random partitions provide a mapping function that attributes all subjects $s \in S_{c}$ to a Hamming embedding $h(s) \in \{\pm1\}^{d}$ where the similarity between two strings indicates the Hamming distance between their images~\cite{jegou2008hamming}.
The chances of any two individuals sharing the very same binary string decreases as the number of regression models $d$ expands. 
Note that the algorithm learns $d$ different \textsc{pls} models assigning feature descriptors obtained from individuals in the positive subset with target value $+1$ whereas those in the negative subset are assigned to target value $-1$.
That is, the assignment of a subject $s$ to the $i$-th positive or negative subset, which is associated with model $m_{i} \in M^{d}$, designates the $i$-th bit in the Hamming embedding.
Each model $m \in M^{d}$ contains a record of the individuals randomly included in its corresponding positive subset.

\vspace{1.5mm}\noindent\textbf{Testing.}
It is worthwhile to mention that the approach is equivalent to estimating Hamming embeddings for all subjects in the gallery set and then comparing them with the binary string generated for query sample $q$.
Given $q$, the algorithm initializes the vote-list histogram and sets all bins (each bin corresponds to a subject $s \in S_{t}$) with zero values. 

When $q$'s feature descriptor is presented to regression models $m \in M^{d}$, they return an array $R^{d}$ composed of $d$ response values $r$ ranging between $-1$ and $+1$.
Each model $m_{i} \in M$ holds its own list of subjects from the gallery set assigned to the $i$-th positive subset.
% This list nominates the bins (classes) that may be incremented by $r_{i}$ in case model $m_{i}$ also satisfies the dictionary criterion.
More precisely, only histogram bins of individuals enlisted in the $i$-th positive subset are incremented by $r_{i}$. % if and only if $m_{i}$ lies on the list of models selected for subject $s$, which is stored in dictionary $D_{[s]}$ . 
In addition, response values $r_{i} \in R$ go through a filtering function $r_{i} = max(0, r_{i})$, which resembles the ReLU activation function.  

We expect that when a \emph{known} probe sample is projected onto the ensemble, the regression response $r_{i} \in R^{d}$ associated with regression model $m_{i} \in M^{d}$ will indicate whether its corresponding class belongs to the $i$-th positive or negative subset.
Particularly, it implies that a probe sample $q$ resembles subjects from the $i$-th positive subset if $r_{i}$ converges towards $+1$ in the testing stage.
If the regression response $r_{i}$ is negative, it suggests that the query sample $q$ belongs to someone enrolled in the $i$-th negative subset.
Then, the algorithm performs no increment in the vote-list histogram since negative responses are filtered out by the ReLU-like function.

\vspace{1mm}\noindent\textbf{Gallery Detection.}
As a query sample $q$ is projected to all regression models $m \in M$ and the vote-list bins are progressively incremented, the proposed method sorts the histogram in descending order in behalf of creating a ranking of candidates.
After the ranking generation, the vote-list histogram is arranged in such a way that individuals with higher probability of matching probe $q$ lie on top of the candidates ranking.

Since we are dealing with an open-set task, it is essential to find out if the gallery set indeed contains an identity that matches $q$.
Therefore, the same list of candidates is explored as the approach computes the ratio of the top scorer to the mean of the next two following individuals, candidates two and three~\cite{vareto2017towards}.
If the ratio is higher than a specified threshold, the method considers the probe sample as a known subject.
If not, the pipeline halts since there is no further reason to continue searching for a subject holding low probability of being previously enrolled in the gallery.

\subsection{Technical Analysis}
% The method designed by Vareto~\etal\cite{vareto2017towards} does not take advantage of the dictionary $D$.
% As a consequence, all models, including inaccurate ones, may end up harming the voting structure since erroneous values can be added to bins of their associated subject identities. 
% It also decrements the vote-list histogram when the response value $r < 0$, resulting in bins with negative values. 

A practical advantage of the proposed algorithm is that the search for faces resembling a given probe image is reduced to a search in the metric embedding space, which is executed quickly. 
The adoption of the vote-list histogram is a manner of weighting the hamming bits when individuals lie in positive subsets.
With clustering, given a dataset of numerous faces and a probe image, the idea is to retrieve identities that are similar to the query without comparing it to every subject (class) enrolled in the gallery set.

The method's underlying structure makes the ensemble size independent from the number of gallery-registered subjects. 
Besides, no retraining is required when extra individuals are inserted into the training set.
With this autonomous characteristic, the combination of \textsc{apc} clusters and \textsc{pls} ensembles is capable of achieving satisfactory results and outperform literature techniques by enclosing far fewer models than traditional multi-class classification schemes~\cite{huang2011extreme}.
Consequently, the designed algorithm provides scalability to galleries composed of numerous individuals and on which regular open-set face recognition methods may probably fail to respond in low computational time~\cite{dos2016partial}.

   \section{Experimental Results}
This section presents the experimental evaluation of the approach described in Section~\ref{sec:method}.
From now on, we refer to the developed method as \textsc{cot} since it consists of a clustering strategy followed by training a collection of \textsc{pls} models.

\subsection{Setup}
We explore the Scikit-Learn library for Python, an efficient open-source tool for data analysis and mining.
\textsc{cot} operates on high-quality deep features, extracted with publicly available \textsc{vggface} network~\cite{cao2018vggface2}, a convolutional neural network designed for face detection and recognition, based on \textsc{vgg16}, \textsc{resnet50} and \textsc{senet50} architectures.
TensorFlow is the adopted neural network library, high-leveled with Keras \textsc{api} for fast experimentation.
All algorithm evaluations are performed on Intel Xeon E5-2630 \textsc{cpu} with 2,30 GHz and 12\textsc{gb} of \textsc{ram} using Ubuntu 18.04 \textsc{lts} operating system, no more than 8\textsc{gb} of \textsc{ram} was required though.

\subsection{Datasets}
Experiments regarding parameter validation and comparison with state-of-the-art methods are conducted with unofficial open-set protocols on \textsc{lfw} and \textsc{ytf} benchmarks\footnote{\textbf{\textsc{fyi}.} It is important to mention that throughout the development process, we have requested datasets designed for open-set face tasks, such as \textsc{ijb-a}~\cite{klare2015pushing}, \textsc{ijb-c}~\cite{maze2018iarpa}, MegaFace~\cite{kemelmacher2016megaface}, and the \textsc{uccs} competition~\cite{gunther2017unconstrained}.
We have contacted their administrators who stated that these benchmarks are either unavailable for copyright issues or erroneous metadata. Consequently, these face datasets are not included in our experimental section.}.

\vspace{0.75mm}\noindent\textbf{\textsc{lfw}~\cite{huang2008labeled}.}
A database of face photographs containing approximately 13,000 uncontrolled face images of almost six thousand individuals. 
% The images were taken in entirely unconstrained situations with non-cooperative individuals.
The original database includes four different  \textsc{lfw} sets images as well as three different types of aligned images.
Following a recent work from literature\cite{gunther2017toward}, we adopt the funneled aligned \textsc{lfw} images.

\vspace{0.75mm}\noindent\textbf{\textsc{ytf}~\cite{wolf2011face}.}
Contains face videos captured for investigating the problem of unconstrained face verification in videos.
It consists of 3,425 videos of 1,595 different people.
Organizers provide the entire dataset broken into frames and preprocessed with facial detection and alignment.

% \vspace{1.0mm}\noindent\textbf{\textsc{fyi}.}
% It is important to mention that throughout the development process, we have requested datasets designed for open-set face tasks, such as \textsc{ijb-a}~\cite{klare2015pushing}, \textsc{ijb-c}~\cite{maze2018iarpa} and the \textsc{uccs} competition~\cite{gunther2017unconstrained}.
% After requesting the benchmarks through the appropriate application forms, we have contacted their administrators who stated that these benchmarks are either unavailable for copyright issues or erroneous metadata.
% Consequently, these face datasets are not included in our experimental section.

% Email 01
% Sorry, but after someone published metadata that made matching much easier we pulled the data.  
% We are working on revising it before our next competition but have no timeline for that since we do not have funding to work on it.
% Terrance Boult

% Email 02
% Hi William:
% We are not allowed to redistribute this dataset.
% I am ccing the IARPA JANUS PM and Technical adviser.
% Rama

\subsection{Protocols}
% There is not a universal agreement when it comes to protocols for open-set face identification.
In this experimental evaluation, we adopt conventions proposed in recent literature works, which are identified by each work's corresponding first author's name:
% ~\cite{liao2014benchmark, gunther2017toward, ferrari2018extended}

\vspace{0.75mm}\noindent\textbf{\textsc{rowden} protocol~\cite{best2014unconstrained}.}
This protocol partitions \textsc{lfw} so that the gallery set consists of 596 identities, in which each individual has at least two training images and exactly a single probe face sample.
The remaining 4,494 face samples comprise the set of impostor probe images.
% Papers using this protocol:
% - Unconstrained Face Recognition: Identifying a Person of Interest From a Media Collection
% - DeepID3: Face Recognition with Very Deep Neural Networks

\vspace{0.75mm}\noindent\textbf{\textsc{g{\"u}nther} protocol~\cite{gunther2017toward}.}
This protocol splits \textsc{lfw} subjects into three disjoint groups: 
602 identities containing more than three face samples compose the \textit{known} subset.
Other 1,070 individuals holding two or three images constitute the \textit{known unknowns} subset (distractors). 
Lastly, 4,096 identities with a single image each only available during test time.
% Papers using this protocol:
% - Toward Open-Set Face Recognition

% Another protocol:
% - A benchmark study of large-scale unconstrained face recognition

\vspace{0.75mm}\noindent\textbf{\textsc{martinez} protocol~\cite{martinez2018toward}.}  
This protocol randomly divides the \textsc{ytf} into ten training and test subsets.
It is characterized by the concept of \textit{openness} (\textsc{op}): defined as the ratio between the genuine comparisons and the impostor comparisons in the probe set.

\subsection{Metrics}
The Receiver Operating Characteristic (\textsc{roc}), its associated Area Under Curve (\textsc{auc}), and the Cumulative Match Characteristic (\textsc{cmc}) are widely employed in closed-set tasks~\cite{jain2000biometric}.
We also incorporate the Detection and Identification Rate (\textsc{dir}) and False Alarm Rate (\textsc{far}). 
% \textsc{dir} is a probability estimate that a subject enrolled in the gallery is detected whereas \textsc{far} estimates the likelihood a non-enrolled individual is characterized as belonging to the gallery set. 
Plotting \textsc{dir} \textit{vs.} \textsc{far} produces a chart known as Open-set \textsc{roc}, a metric generally used to evaluate approaches composed by filtering and identification steps~\cite{jain2011handbook}.

\subsection{Parameter Selection}
Experiments carried out in this subsection contemplate \textsc{vggface}'s \textsc{resnet50} architecture and \textsc{lfw} dataset in a closed-set fashion.
% More precisely, we select the parameters considering only known identities, avoiding tests that include face samples from unknown subjects. 
We follow both \textsc{rowden} and \textsc{g{\"u}nther} protocols to a certain extent: a random dataset partition, selecting 602 subjects to compose gallery and probe sets as well as setting the maximum of three samples for training and a single sample for testing.

\vspace{1.5mm}\noindent\textbf{\textsc{pls} dimensions.}
\textsc{pls} regression models only require a single parameter: the number of components in the latent space.
Table~\ref{tab:plscomp} presents \textsc{cmc} results as we gradually increase the number of \textsc{pls} components from 5 to 20, but keep fixed ensemble size $d$ of 50 \textsc{pls} models.
Since there is a minor advantage in using a particular dimension size, we set the number of \textsc{pls} components $c$ to 10 for all following experiments after 30 executions.

\begin{table}[h]
    \small
    \centering
    \begin{tabular}{rr|ccccccc} \hline
        \multicolumn{2}{l}{\textsc{pls} dimensions}  &    5 &    7 &   10 &   12 &   14 &   20 \\ \hline \hline
        \multirow{2}{*}{Rank-01} & \textsc{avg}      & 93.0 & 94.4 & 96.7 & 96.2 & 94.3 & 93.5 \\ 
                                 & \textsc{std}      & 2.02 & 1.67 & 1.62 & 1.66 & 1.46 & 1.40 \\ \hline
        \multirow{2}{*}{Rank-05} & \textsc{avg}      & 94.8 & 96.0 & 98.1 & 98.3 & 96.4 & 95.7 \\
                                 & \textsc{std}      & 1.25 & 1.06 & 1.03 & 0.99 & 1.10 & 1.15 \\ \hline
        \multirow{2}{*}{Rank-10} & \textsc{avg}      & 95.8 & 96.2 & 98.9 & 98.9 & 97.8 & 96.7 \\
                                 & \textsc{std}      & 1.07 & 1.05 & 0.93 & 1.00 & 0.83 & 0.86 \\ \hline
    \end{tabular}
    \vspace{-2.0mm}
    \caption{Variable number of \textsc{pls} dimensions and their corresponding \textsc{cmc} averaged results (\%) on different \textit{Ranks}.}
    \label{tab:plscomp}
    \vspace{-2.0mm}
\end{table}

\vspace{1.0mm}\noindent\textbf{\textsc{pls} models.}
% To check how \textsc{cot} responds to variable ensemble sizes, we analyze the behavior of the proposed approaches by varying the number of \textsc{pls} models.
Progressively adding regression models results in extra random partitions, further models training and extra probe feature projections during evaluation time.
Initially, a small embedding size $d$ does not seem adequate for discerning when an individual is registered in the gallery set and, consequently, retrieve the correct identity.

\begin{table}[h]
    \small
    \centering
    \begin{tabular}{rr|ccccccc} \hline
        \multicolumn{2}{l}{\textsc{pls} models} &   30 &  40 &  50 &   60 &   80 &    100 \\ \hline \hline
        \multirow{2}{*}{Rank-01} & \textsc{avg} & 93.0 & 95.4 & 96.7 & 97.2 & 97.3 & 97.6 \\ 
                                 & \textsc{std} & 2.40 & 1.64 & 1.62 & 1.15 & 1.14 & 1.03 \\ \hline
        \multirow{2}{*}{Rank-05} & \textsc{avg} & 95.3 & 97.3 & 98.1 & 98.6 & 98.9 & 99.0 \\
                                 & \textsc{std} & 1.99 & 1.23 & 1.03 & 0.66 & 0.43 & 0.35 \\ \hline
        \multirow{2}{*}{Rank-10} & \textsc{avg} & 97.9 & 98.6 & 98.9 & 99.1 & 99.2 & 99.2 \\
                                 & \textsc{std} & 1.38 & 1.12 & 0.93 & 0.42 & 0.29 & 0.12 \\ \hline
    \end{tabular}
    \vspace{-2.0mm}
    \caption{Variable number of \textsc{pls} models and their corresponding \textsc{cmc} averaged results (\%) on different \textit{Ranks}.}
    \label{tab:plsmodels}
    \vspace{-2.0mm}
\end{table}

Table \ref{tab:plsmodels} exposes how the number of \textsc{pls} models affects the implemented method.
There is a meaningful improvement when varying the ensemble size from 30 to 100.
However, we observe no significant accuracy increase when the number of regression models exceeds 50 learning algorithms.
This experiment advocates the claim in which we declare that the ensemble size is not directly dependent on the number of known individuals.
Therefore, we fix the number of \textsc{pls} models on 60 in the subsequent experiments.
We are inclined to believe that either a reduced number of samples per identity or an expanded number of individuals enrolled in the gallery set during training time may require additional \textsc{pls} regression models in order to keep \textsc{cmc} \textit{Rank-1} values high.

\vspace{1.0mm}\noindent\textbf{Potential Accuracy.}
The Maximum Achievable Recognition Rate (\textsc{marr})~\cite{dos2016partial} permits the evaluation of the filtering approach performance, adopted to reduce the \textsc{pls} ensemble size.
It demonstrates how an error originated during the picking of best clusters propagates to the rest of the pipeline.

\begin{figure}[!t]
    \centering
    \vspace{-2.0mm}
    \includegraphics[trim={0 0.5cm 0 0},clip,width=0.90\columnwidth]{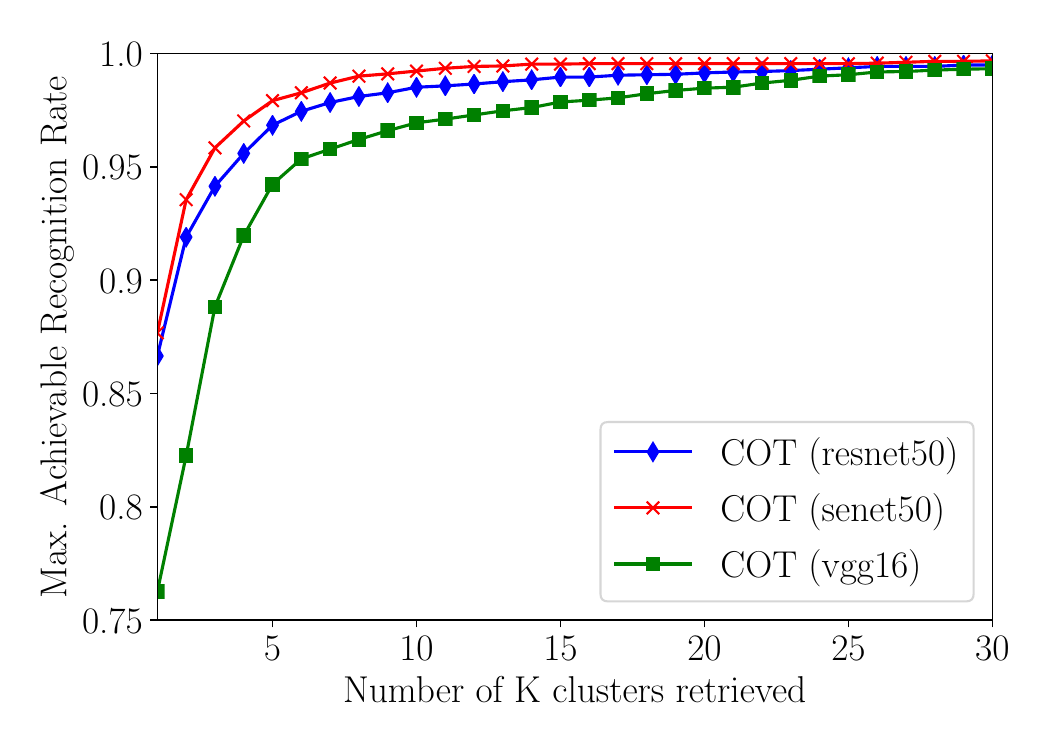}
    \caption{Comparison of the number of clusters retrieved and the maximum achievable recognition rate.}
    \label{fig:marr}
    \vspace{-4.0mm}
\end{figure}

\textsc{marr} assumes that a flawless face identification method is employed and, therefore, indicates the upper bound rate for both filtering and identification pipeline.
Failing to select the best $k$ clusters results in trouble returning the correct identity for a given probe image.
Figure~\ref{fig:marr} demonstrates that an insufficient number of retrieved clusters severely diminishes the approach's capacity.
However, the ensemble may be able to fix the inferiority of \textsc{vgg16} with larger numbers of $k$.
Based on the experiments, we set $k$ to 20 in all experiments described in the following section.

% Based on the chart, the ensemble of \textsc{pls} may be able to fix the inferiority of \textsc{vgg16} with larger numbers of $k$.

\subsection{Literature Comparison}
On the contrary of previous experiments that only deal with known individuals, this subsection focuses on assessing the complete open-set identification pipeline by contrasting the proposed \textsc{cot} approach with state-of-the-art works.
Besides, we incorporate both \textsc{vggface}'s architectures: \textsc{resnet50} and \textsc{senet50}. 

\begin{figure*}[!ht]
    \centering
    \begin{subfigure}[b]{0.49\textwidth}
        \centering
        \includegraphics[width=0.92\textwidth]{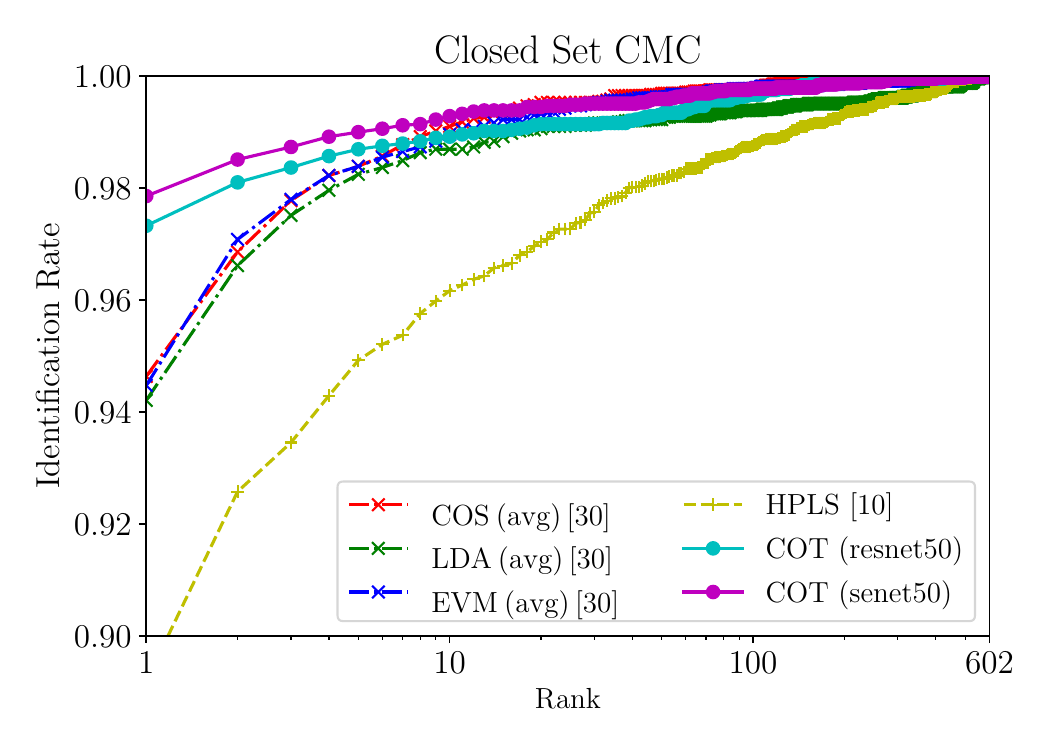}
    \end{subfigure}
    \hfill
    \begin{subfigure}[b]{0.49\textwidth}
        \centering
        \includegraphics[width=0.92\textwidth]{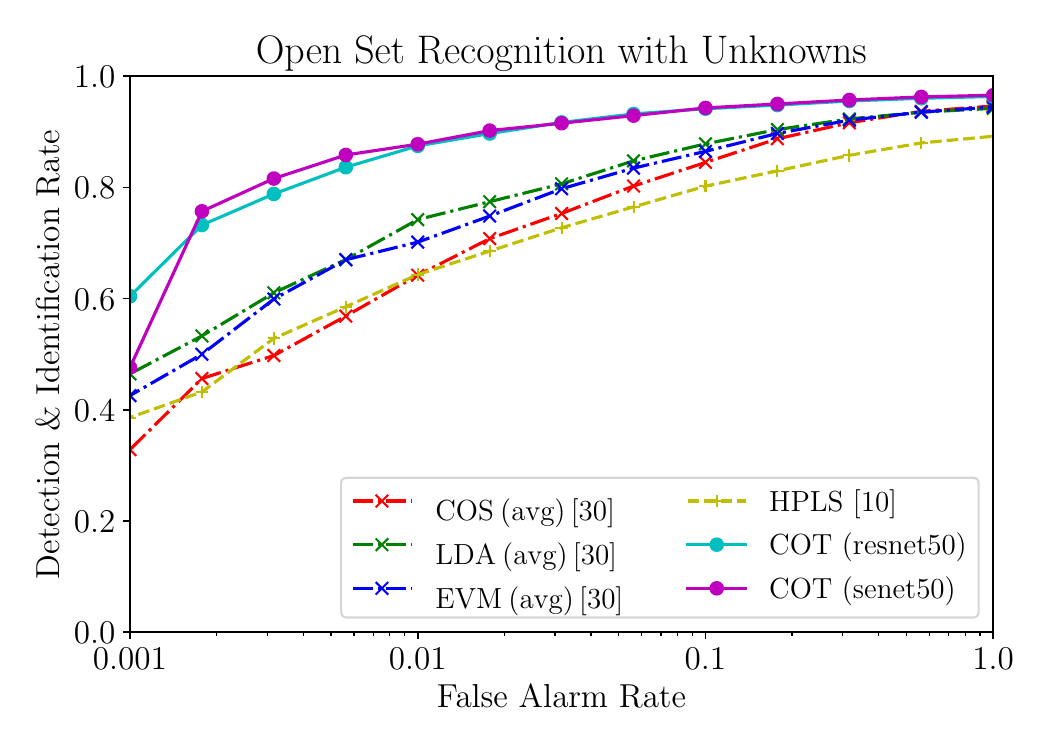}
    \end{subfigure}
    \vspace{-5.0mm}
    \caption{\textsc{cmc} and open-set \textsc{dir} comparison between methods following \textsc{g{\"u}nther} protocol on \textsc{lfw}.}
    % \caption{Comparison between methods following \textsc{g{\"u}nther} protocol on \textsc{lfw} dataset: \textsc{cmc} and open-set \textsc{dir} curves are given for all six different methods.}
    \label{fig:gunther_protocol}
    \vspace{-2.0mm}
\end{figure*}

Figure~\ref{fig:gunther_protocol} presents the comparison between literature methods in total accordance with \textsc{g{\"u}nther} protocol.
Note that \textsc{vggface} architectures embedded in \textsc{cot} share similar performance nature.
The \textsc{cmc} chart demonstrates a superior performance of \textsc{cot} up to \textit{Rank-10}.
The proposed method outperformed all literature works when evaluating the open-set \textsc{roc} curve, displayed in the chart on the right.
The \textsc{dir}~\textit{vs.}~\textsc{far} plot evaluates both types of \textit{unknowns}, which consists of the average evaluated methods' performance on \textit{known unknowns} and \textit{unknown unknowns}.

\begin{table}[!ht]
    \small
    \centering
    \begin{tabular}{r|cccc} \hline
        Method            & \textsc{DeepID3} & \textsc{DeepID2} & \textsc{Hpls} & \textsc{cot\textsubscript{res50}} \\ \hline \hline 
        Rank-1            &             96.0 &             95.0 &          90.7 &                              98.1 \\
        \textsc{dir@0.01} &             81.4 &             80.7 &          64.3 &                              82.3 \\ \hline
    \end{tabular}
    \vspace{-2.0mm}
    \caption{Comparison between methods following \textsc{rowden} protocol on \textsc{lfw} dataset.}
    % \caption{Comparison between methods following \textsc{rowden} protocol on \textsc{lfw} dataset: \textsc{cmc} and open-set \textsc{dir} values are given for four different methods.}
    \label{tab:rowden_protocol}
    \vspace{-5.0mm}
\end{table}

In Figure~\ref{fig:gunther_protocol}, approximately 88\% of known subjects are correctly identified with our proposed method.
An open-set \textsc{roc} curves's top left corner indicates the optimal point. 
For a clear understanding of the results, a false alarm rate of 0.01 implies that one out of 100 unknown individuals are mistakenly assigned to an identity from the face gallery\footnote{As an illustration, think of a biometric surveillance system at a football stadium that takes around 100 face pictures in a minute. The system is going to trigger one false alarm every sixty seconds as it searches for each one of them in a mugshot database. On the other hand, \textsc{cot} would successfully recognize nearly 90\% of lawbreakers and fugitives.}.
Table~\ref{tab:rowden_protocol} compares closed-set \textit{Rank-1} and open-set \textsc{dir} identification rates when the number of false alarms (\textsc{far}) is fixed on 1\% under \textsc{rowden} protocol.  
\textsc{cot\textsubscript{resnet50}} algorithm achieved best \textsc{cmc}@\textit{Rank-1} accuracy of 98.1 as well as a \textsc{dir@far=0.01} of 82.3.
We expected a better performance of the proposed method in this \textsc{rowden} protocol as it provides more samples per class in the training set.

As mentioned in Sections~\ref{sec:introduction}~and~\ref{sec:method}, \textsc{cot} is not directly dependent on the number of classes, but to the number of subjects encompassed by the top $k$ clusters.
Both \textsc{lfw} protocols provide over 590 subject identities during training time yet the proposed method did not require an ensemble with more than 60 \textsc{pls} regression models.
In addition, since the random partitions generate balanced data splits, we can replace \textsc{pls} by other long-established learning algorithms, such as support vector machines~\cite{steinwart2008support} and neural networks, without facing instability issues that may cause accuracy drop due to disproportionate training data.

\begin{table}[!ht]
    \small
    \centering
    \begin{tabular}{r|ccc} \hline
        Method            & \textsc{ShuffleNet} & \textsc{ShuffleFace} & \textsc{cot\textsubscript{res50}} \\ \hline \hline 
        \textsc{op}=$0.2$ &               82.65 &                86.83 &                             87.65 \\
        \textsc{op}=$0.5$ &               79.22 &                85.52 &                             83.79 \\
        \textsc{op}=$0.9$ &               78.03 &                84.61 &                             79.54 \\ \hline
    \end{tabular}
    \vspace{-2.0mm}
    \caption{Comparison between methods on \textsc{ytf} dataset.}
    \label{tab:martinez_protocol}
    \vspace{-5.0mm}
\end{table}

Table~\ref{tab:martinez_protocol} presents closed-set \textsc{cmc} results on the \textsc{ytf} database following \textsc{martinez} closed-set protocol (video-to-video).
The proposed \textsc{cot\textsubscript{resnet50}} algorithm achieved best \textsc{cmc}@\textit{Rank-1} accuracy of 87.65 when openness (\textsc{op}) was set to 0.2.
However, as the gallery set increased, \textsc{ShuffleFace}~\cite{martindez2019shufflefacenet} obtained the best results.
% The proposed method was initially designed for open-set face tasks. 
% However, it's generalization capability also attains good results in closed-set problems.
In summary, the proposed method's generalization capability allows it to obtain good results in closed-set problems even though it was initially designed for open-set face tasks.
   \section{Conclusion}

This work details a scalable open-set face identification approach to galleries with hundreds and thousands of subjects. 
The proposed algorithm combines a clustering algorithm with an ensemble of regression models, so that it either retrieves individuals from the gallery set that present substantial similarity to the probe image or considers them unknown subjects.

We evaluated different parameters to check how they would impact the proposed algorithm.
All experiments were conducted on \textsc{lfw} and \textsc{ytf}, face databases initially proposed for face verification.
We adopted third-party protocols with the intent of establishing appropriate comparisons with state-of-the-art methods. 
Both databases do not provide enough number of individuals for a pertinent study regarding scalability towards large face galleries.
In summary, open-set face identification remains a difficult problem and more effort is necessary to make it fully effective.

\ackfootnote{
    \textbf{Acknowledgments.} The authors would like to thank the National Council for Scientific and Technological Development -- CNPq (Grants~438629/2018-3 and~309953/2019-7), the Minas Gerais Research Foundation -- FAPEMIG (Grants~APQ-00567-14 and~PPM-00540-17), the Coordination for the Improvement of Higher Education Personnel -- CAPES (DeepEyes Project). This study was financed in part by the Coordena\c{c}\~{a}o de Aperfei\c{c}oamento de Pessoal de N\'{i}vel Superior - Brasil (CAPES) - Finance Code 001.
}

   {  
      \footnotesize
      \bibliographystyle{unsrt}
      \bibliography{bibliography}
   }

\end{document}